\documentclass{article}
\usepackage[utf8]{inputenc}

\usepackage{graphicx}
\usepackage{amssymb,amsmath,mathtools,amscd}
\usepackage{amsthm}
\usepackage{mathabx}
\usepackage{epstopdf}
\usepackage{tikz}
\usepackage{tikz-cd}
\usepackage{stmaryrd}
\usepackage{xcolor}

\usepackage{algpseudocode}
\usepackage{algorithm}
\newcommand{\CC}{\mathbb{C}}
\newcommand{\RR}{\mathbb{R}}

\newcommand{\Maps}{\operatorname{Maps}}

\theoremstyle{definition}

\usepackage{xfrac}

\title{Machine learning and invariant theory}
\author{Ben Blum-Smith and Soledad Villar}
\date{\small Department of Applied Mathematics and Statistics \\ Mathematical Institute for Data Science \\ Johns Hopkins University}

\usepackage{comment}

\begin{document}
\maketitle
\begin{abstract}
\noindent
      Inspired by constraints from physical law, equivariant machine learning restricts the learning to a hypothesis class where all the functions are equivariant with respect to some group action. Irreducible representations or invariant theory are typically used to parameterize the space of such functions. 
   In this article, we introduce the topic and explain a couple of methods to explicitly parameterize equivariant functions that are being used in machine learning applications.
   In particular, we explicate a general procedure, attributed to Malgrange, to express all polynomial maps between linear spaces that are equivariant under the action of a group $G$, given a characterization of the invariant polynomials on a bigger space. The method also parametrizes smooth equivariant maps in the case that $G$ is a compact Lie group.

\end{abstract}

\section{Introduction}

Modern machine learning has not only surpassed the state of the art in many engineering and scientific problems, but also it has had an impact on society at large, and will likely continue to do so. This includes deep learning, large language models, diffusion models, etc. In this article, we give an account of certain mathematical principles that are used in the definition of some of these machine learning models, and we explain how classical invariant theory plays a role in them. 

In supervised machine learning, we typically have a training set $(x_i,y_i)_{i=1}^n$, where $x_i\in \mathbb R^d$ are the data points and $y_i \in \mathbb R^k$ are the labels. A typical example is image recognition, where the $x_i$ are images and the $y_i$ are image labels (say, `cat' or `dog'), encoded as vectors. The goal is to find a function $\hat f$ in a hypothesis space $\mathcal F$, that not only approximately interpolates the training data ($\hat f(x_i)\approx y_i$), but also performs well on unseen (or held-out) data. The function $\hat f$ is called the {\em trained model}, {\em predictor}, or {\em estimator}. 
In practice, one parametrizes the class of functions $\mathcal F$ with some parameters $\theta$ varying over a space $\Theta$ of parameter values sitting inside some $\mathbb R^s$; in other words, $\mathcal F = \{f_\theta : \mathbb R^d \to \mathbb R^k,\; \theta \in \Theta \subseteq \mathbb R^s \}$. Then one uses local optimization (in $\theta$) to find a function in $\mathcal F$ that locally and approximately minimizes a prespecified empirical loss function $\ell$ which compares a candidate function $f_\theta$'s values on the $x_i$ with the ``true" target values $y_i$. In other words, one approximately solves $\theta^* := \operatorname{argmin}_\theta \sum_{i=1}^n \ell (f_\theta (x_i) , y_i)$, and then takes $\hat f=f_{\theta^*}$.

Modern machine learning performs regressions on classes of functions that are typically overparameterized (the dimension $s$ of the space of parameters is much larger than the number $n$ of training samples), and in many cases, several functions in the hypothesis class $\mathcal F$ can interpolate the data perfectly (they can even interpolate nonsensical data \cite{zhang2017understanding}!). Moreover, the optimization problem is typically non-convex. Therefore the model performance is highly dependent on how the class of functions is parameterized and the optimization algorithms employed. 

The parameterization of the hypothesis class of functions is what in deep learning is typically referred to as \emph{the architecture}. In recent years, the most successful architectures have been ones that use properties or heuristics regarding the structure of the data (and the problem) to design the class of functions: convolutional neural networks for images, recurrent neural networks for time series, graph neural networks for graph-structured data, transformers, etc. Many of these design choices are related to the symmetries of the problem: for instance, convolutional neural networks can be translation equivariant, and transformers can be permutation invariant. 

When the learning problem comes from the physical sciences, there are concrete sets of rules that the function being modeled must obey, and these rules often entail symmetries. The rules (and symmetries) typically come from coordinate freedoms and conservation laws \cite{villar2023passive}. 
One classical example of these coordinate freedoms is the scaling symmetry that comes from dimensional analysis \cite{villar2022dimensionless, bakarji2022dimensionally} (for instance, if the input data to the model is rescaled to change everything that has units of kilograms to pounds, the predictions should scale accordingly). 
In order to do machine learning on physical systems, researchers have designed models that are consistent with physical law; this is the case for physics-informed machine learning \cite{karniadakis2021physics}, neural ODEs and PDEs \cite{chen2018neural, hamiltonian}, and equivariant machine learning \cite{zaheer2017deep, cohen2016group, cohen2019gauge, maron2018invariant, weiler2021coordinate}. 

Given data spaces $V,W$ and a group $G$ acting on both of them, a function $f:V\to W$ is equivariant if $f(g\cdot v) = g\cdot f(v)$ for all $g\in G$ and all $v\in V$. Many physical problems are equivariant with respect to rotations, permutations, or scalings. For instance, consider a problem where one uses data to predict the dynamics of a folding protein or uses simulated data to emulate the dynamics of a turbulent fluid. Equivariant machine learning restricts the hypothesis space to a class of equivariant functions. The philosophy is that every function that the machine learning model can express is equivariant, and therefore consistent with physical law. 

Symmetries were used for machine learning (and in particular neural networks) in early works \cite{shawe1989building, shawe1993symmetries, wood1996representation}, and more recently they have been revisited in the context of deep learning. There are three main ways to implement symmetries. The simplest one, for discrete groups, is parameter sharing \cite{ravanbakhsh2017sharing, cohen2016group, zaheer2017deepsets}, where a parameterization of invariant and equivariant functions can be found by averaging arbitrary functions.  The second approach, explained in the next section, uses classical representation theory to parameterize the space of equivariant functions \cite{kondor2018n, thomas2018tensor, fuchs2020se, geiger2022e3nn}. The third approach, the main point of this article, uses invariant theory.

As an example, we briefly discuss graph neural networks (GNNs), which have been a very popular area of research in the past couple of years. GNNs can be seen as equivariant functions that take a graph represented by its adjacency matrix $A \in \mathbb R^{n\times n}$ and possible node features $X \in \mathbb R^{n\times n}$, and output an embedding $f(A, X) \in \mathbb R^{n\times d}$ so that $f(\Pi A \Pi^\top, \Pi X) = \Pi f(A, X)$ for all $\Pi$ $n\times n$ permutation matrices \cite{duvenaud2015convolutional, bruna2013spectral, gilmer2017neural, maron2018invariant, chen2019equivalence}. Graph neural networks are typically implemented as variants of graph convolutions or message passing, which are equivariant by definition.
However, many equivariant functions cannot be expressed with these architectures. Several recent works analyze the expressive power of different GNN architectures in connection to the graph isomorphism problem. 

Beyond graphs, equivariant machine learning models have been extremely successful at predicting molecular structures and dynamics \cite{batzner20223,miller2020relevance,thiede2021autobahn}, protein folding \cite{jumper2021highly}, protein binding \cite{stark2022equibind},  and simulating turbulence and climate effects \cite{yu-physics, wang2021incorporating, wang2022approximately}. Theoretical developments have shown the universality of certain equivariant models \cite{dym2020universality, maron2019universality, bokman2022zz, yarotsky2018universal}, generalization improvements of equivariant machine learning models over non-equivariant baselines \cite{bietti2021sample, elesedy2021kernel, elesedy2021provably, mei2021learning}, and there has been some recent work studying the inductive bias of equivariant machine learning \cite{lawrence2021implicit}, and its relationship with data augmentation \cite{wang2022data,gerken2022equivariance}.

\section{Equivariant convolutions and multi-layer perceptrons} \label{sec.equiv}

Modern deep learning models have evolved from the classical artificial neural network perceptron \cite{rosenblatt1958perceptron}. The multi-layer perceptron model takes an input $x$ and outputs $F(x)$ defined to be the composition of affine linear maps and non-linear entry-wise functions. Namely,
\begin{equation}
    \label{eq.gnn.eq}
F(x)= \rho \circ   L_T \circ \ldots.\, \circ   L_2 \circ \rho \circ   L_1 (x)~,
\end{equation}
where $\rho$ is the (fixed) entry-wise non-linear function and $L_i: \mathbb R^{d_i}\to \mathbb R^{d_{i+1}}$ are affine linear maps to be learned from the data. The linear maps $L_i$ can be expressed as $L_i(x) = A_i x + b_i$ where $A_i \in \mathbb R^{d_i\times d_{i+1}}$ and $b_i \in \mathbb R^{d_{i+1}}$. In this example each function $F$ is defined by the parameters $\theta=(A_i,b_i)_{i=1}^T$. 

The first neural network that was explicitly equivariant with respect to a group action is the convolutional neural network \cite{lecun1989backpropagation}. The observation is that if the $N\times N$ input images $x$ are seen in the torus $\sfrac{\mathbb Z}{N\mathbb Z} \times \sfrac{\mathbb Z}{N\mathbb Z}$, the linear equivariant maps are cross-correlations (which in ML are referred to as convolutions) with fixed filters. 
The idea of restricting the linear maps to satisfy symmetry constraints was generalized to equivariance with respect to discrete rotations and translations in \cite{wood1996representation, cohen2016group}, and to general homogenous spaces in \cite{kondor2018clebsch, cohen2019general}. Note that when working with arbitrary groups there are restrictions on the functions $\rho$ for the model to be equivariant.

Classical results in neural networks show that multi-layer perceptrons can universally approximate any continuous function \cite{barron1993universal}. However, that is not true in general in the equivariant case. Namely, functions expressed as \eqref{eq.gnn.eq} where the $L_i$ are linear and equivariant may not universally approximate all continuous equivariant functions. In some cases, there may not even exist non-trivial linear equivariant maps. 

One popular idea to address this issue is to extend this model to use equivariant linear maps on tensors. Now  $L_i: \mathbb R^{d^{\otimes k_i}}\to \mathbb R^{d^{\otimes k_{i+1}}}$ are linear equivariant maps (where the action in the tensor product is defined as the tensor product of the action in each component and extended linearly). Now the question is how can we parameterize the space of such functions to do machine learning? The answer is via Schur's lemma. 

A representation of a group $G$ is a map $\phi: G\to \operatorname{GL}(V)$  that satisfies $\phi(g_1g_2)=\phi(g_1)\phi(g_2)$ (where $V$ is a vector space and $\operatorname{GL}(V)$, as usual, denotes the automorphisms of $V$, that is, invertible linear maps $V\to V$).
A group action of $G$ on $\mathbb R^d$ (written as $\cdot$) is equivalent to the group representation $\phi: G \to \operatorname{GL}(\mathbb R^d)$ such that $\phi(g)(v)=g\cdot v$.
We extend the action $\cdot$ to the tensor product $(\mathbb R^d)^{\otimes k}$ so that the group acts independently in every tensor factor (i.e., in every dimension), namely $\phi_k = \otimes_{r=1}^{k} \phi: G \to \operatorname{GL}((\mathbb R^d)^{\otimes k})$. 

The first step is to note that a linear equivariant map $  L_i: (\mathbb R^d)^{\otimes k_i} \to (\mathbb R^d)^{\otimes k_{i+1}}$ corresponds to a map between group representations such that $  L_i \circ \phi_{k_i}(g) = \phi_{k_{i+1}}(g)\circ   L_i$ for all $g\in G$.
Homomorphisms between group representations are easily parametrizable if we decompose the representations in terms of irreducible representations (aka irreps): 
\begin{align}
\phi_{k_i} &= \bigoplus_{\ell=1}^{T_{k_i}} \mathcal T_{\ell} ~.
\end{align}
In particular, Schur's Lemma says that a map between two irreps over $\mathbb C$ is zero (if they are not isomorphic) or a multiple of the identity (if they are). 

The equivariant neural-network approach consists in decomposing the group representations in terms of irreps and explicitly parameterizing the maps \cite{kondor2018n, thomas2018tensor, fuchs2020se}.
In general, it is not obvious how to decompose an arbitrary group representation into irreps. However in the case where $G=\text{SO}(3)$, the decomposition of a tensor representation as a sum of irreps is given by the Clebsh-Gordan decomposition: \begin{equation}
 {\otimes_{s=1}^k \phi_s}= \oplus_{\ell=1}^{T} \mathcal T_{\ell} \label{eq.decomposition}
\end{equation}
The Clebsh-Gordan decomposition not only gives the decomposition of the RHS of \eqref{eq.decomposition} but also it gives the explicit change of coordinates. This decomposition is fundamental for implementing the equivariant 3D point-cloud methods defined in \cite{fuchs2020se, thomas2018tensor, bogatskiy2020lorentz}. 
Moreover, recent work \cite{dym2020universality} shows that the classes of functions defined in \cite{fuchs2020se, thomas2018tensor} are universal, meaning that every continuous SO(3)-equivariant function can be approximated uniformly in compacts sets by those neural networks. 
However, there exists a clear limitation to this approach: Even though decompositions into irreps are broadly studied in mathematics (a.k.a. plethysm), the explicit transformation that allows us to write the decomposition of tensor representations into irreps is a hard problem in general. It is called the {\em Clebsch-Gordan problem}. There is exciting, recent progress on this problem for large classes of groups \cite{alex2011numerical, ibort2017new}.

\section{Invariant theory for machine learning}\label{sec:inv-thy-for-ml}

An alternative but related approach to the linear equivariant layers described above is the approach based on invariant theory, the focus of this article. Invariant polynomials have been used to design machine learning models \cite{haddadin2021invariant, gripaios2021lorentz, villar2021scalars, villar2022dimensionless, yao2021simple}, which have been applied to cosmology \cite{thiele2022predicting} and molecular dynamics \cite{gasteiger2021gemnet}. This is also implicitly done in \cite{satorras2021n, lim2022sign}. 
In particular, the authors of this note and collaborators \cite{villar2021scalars} explain that for some physically relevant groups---the orthogonal group, the special orthogonal group, and the Lorentz group---one can use classical invariant theory to design universally expressive equivariant machine learning models that are expressed in terms of the generators of the algebra of invariant polynomials. Following an idea attributed to B. Malgrange (that we learned from G. Schwarz), it is shown how to use the generators of the algebra of invariant polynomials to produce a parameterization of equivariant functions for a specific set of groups and actions.

To illustrate, let us focus on $\mathrm{O}(d)$-equivariant functions, namely functions $f:(\mathbb R^d)^n \to \mathbb R^d$ such that $f(Qv_1, \ldots, Qv_n) = Q f(v_1, \ldots, v_n)$ for all $Q\in \mathrm{O}(d)$ and all $v_1,\ldots, v_n \in \mathbb R^d$ (for instance, the prediction of the position and velocity of the center of mass of a particle system). The method of B. Malgrange (explicated below) leads to the conclusion that all such functions can be expressed as 
\begin{equation}\label{eq:span-1}
    f(v_1,\ldots, v_n) = \sum_{j=1}^n f_j(v_1, \ldots, v_n) v_j,
\end{equation}
where $f_j:(\mathbb R^d)^n \to \mathbb R$ are $\mathrm{O}(d)$-invariant functions. Classical invariant theory shows that $f_j$ is $\mathrm{O}(d)$-invariant if and only if it is a function of the pairwise inner products $(v_i^\top v_j)_{i,j=1}^n$. So, in actuality, \eqref{eq:span-1} can be rewritten
\begin{equation}\label{eq:span-the-module}
    f(v_1,\ldots, v_n) = \sum_{j=1}^n p_j\left((v_i^\top v_j)_{i,j=1}^n\right) v_j.
\end{equation}
In other words, the pairwise inner products generate the algebra of invariant polynomials for this action, and every equivariant map is a linear combination of the vectors themselves with coefficients in this algebra.

In this article, we explicate the method of B. Malgrange in full generality, showing how to convert knowledge of the algebra of invariant polynomials into a characterization of the equivariant polynomial (or smooth) maps. 
In Section \ref{big.picture} we explain the general philosophy of the method, and in Section \ref{mechanics} we give the precise algebraic development, formulated as an algorithm to produce parametrizations of equivariant maps given adequate knowledge of the underlying invariant theory. In Section \ref{sec:examples}, we work through several examples.

We note that, for machine learning purposes, it is not critical that the functions $p_j$ are defined on invariant {\em polynomials}, nor that they themselves are polynomials. In the following, we focus on polynomials because the ideas were developed in the context of invariant theory; the arguments explicated below are set in this classical context. However, in \cite{villar2021scalars}, the idea is to pre-process the data by converting the tuple $(v_j)_{j=1}^n$ to the tuple of dot products $(v_i^\top v_j)_{i,j=1}^n$, and then treat the latter as input to the $p_j$'s, which are then learned using a machine learning architecture of one's choice. Therefore, the $p_j$'s are not polynomials but belong to whatever function class is output by the chosen architecture. Meanwhile, some recent works \cite{cahill2022group, dym2022low,olver2023invariants} have proposed alternative classes of separating invariants that can be used in place of the classical algebra generators as input to the $p_j$'s, and may have better numerical stability properties. This is a promising research direction.

\section{Big picture} \label{big.picture}

We are given a group $G$, and finite-dimensional linear $G$-representations $V$ and $W$ over a field $k$. (We can take $k=\RR$ or $\CC$.) We want to understand the equivariant polynomial maps $V\rightarrow W$. We assume we have a way to understand $G$-{\em invariant} polynomials on spaces related to $V$ and $W$, and the goal is to leverage that knowledge to understand the equivariant maps.

The following is a philosophical discussion, essentially to answer the question: why should it be possible to do this? It is not precise; its purpose is just to guide thinking. Below in Section \ref{mechanics} we show how to actually compute the equivariant polynomials $V\rightarrow W$ given adequate knowledge of the invariants. That section is rigorous.

The first observation is that any reasonable family of maps $V\rightarrow W$ (for example linear, polynomial, smooth, continuous, etc.) has a natural $G$-action induced from the actions on $V$ and $W$, and that the $G$-equivariant maps in such a family are precisely the fixed points of this action, as we now explain. This observation is a standard insight in representation and invariant theory.

Let $\Maps(V,W)$ be the set of maps of whatever kind, and let $GL(V)$ (respectively $GL(W)$) be the group of linear invertible maps from $V$ (respectively $W$) to itself. Given $f\in \Maps(V,W)$ and $g\in G$ and $v\in V$, we define the map $gf$ by
\begin{equation}\label{eq.G-action-on-maps}
gf := \psi(g)\circ f \circ \phi(g^{-1}),
\end{equation}
where $\phi: G \rightarrow GL(V)$ and $\psi: G\rightarrow GL(W)$ are the group homomorphisms defining the representations $V$ and $W$. The algebraic manipulation to verify that this is really a group action is routine and not that illuminating. A perhaps more transparent way to understand this definition of the action as ``the right one" is that it is precisely the formula needed to make this square commute:

\begin{center}
\begin{tikzcd}
V \arrow[r, "f"] \arrow[d, "\phi(g)"] & W \arrow[d, "\psi(g)"] \\
V \arrow[r, "gf"] & W
\end{tikzcd}
\end{center}
It follows from the definition of this action that the condition $gf = f$ is equivalent to the statement that $f$ is $G$-equivariant. The square above automatically commutes, so $gf=f$ is the same as saying that the below square commutes---
\begin{center}
\begin{tikzcd}
V \arrow[r, "f"] \arrow[d, "\phi(g)"] & W \arrow[d, "\psi(g)"] \\
V \arrow[r, "f"] & W
\end{tikzcd}
\end{center}
---and this is what it means to be equivariant.

An important special case of \eqref{eq.G-action-on-maps} is the action of $G$ on $V^*$, the linear dual of $V$. This is the case $W=k$ with trivial action, and \eqref{eq.G-action-on-maps} reduces to $g\ell := \ell\circ \phi(g^{-1})$. This is known as the {\em contragredient action}. We will utilize it momentarily with $W$ in the place of $V$.

The second observation is that $\Maps(V,W)$ can be identified with functions from a bigger space to the underlying field $k$ by ``currying", and this change in point of view preserves the group action. Again, this is a standard maneuver in algebra. Specifically, given any map $f\in \Maps(V,W)$, we obtain a function $\tilde f:V\times W^* \rightarrow k$, defined by the formula
\begin{align*}
\tilde f : V\times W^* &\rightarrow k \\
(v,\ell) &\mapsto \ell(f(v)).
\end{align*}
Note that the function $\tilde f$ is linear homogeneous in $\ell\in W^*$. Conversely, given any function $f' : V\times W^* \rightarrow k$ that is linear homogeneous in the second coordinate, we can recover a map $f:V\rightarrow W$ such that $f' = \tilde f$, by taking $f(v)$ to be the element of $W$ identified along the canonical isomorphism $W\rightarrow W^{**}$ with the functional on $W^*$ that sends $\ell\in W^*$ to $f'(v,\ell)$---this functional is guaranteed to exist by the fact that $f'$ is linear homogeneous in the second coordinate. An observation we will exploit in the next section is that the desired functional is actually the total derivative of $f'(v,\ell)$ with respect to $\ell$.

This construction gives an identification of $\Maps(V,W)$ with a subset of $\Maps(V\times W^*,k)$. Furthermore, there is a natural action of $G$ on $\Maps(V\times W^*,k)$, defined precisely by the above formula \eqref{eq.G-action-on-maps} with $V\times W^*$ in place of $V$, $k$ in place of $W$, and trivial action on $k$;\footnote{The action of $G$ on $V\times W^*$ is defined by acting separately on each factor; the action on $W^*$ is the contragredient representation defined above.} and the identification described here preserves this action. Therefore, the fixed points for the $G$-action on $\Maps(V,W)$ correspond with fixed points for the $G$-action on $\Maps(V\times W^*,k)$, which are {\em invariant functions} (since the action of $G$ on $k$ is trivial).

What has been achieved is the reinterpretation of equivariant maps $f\in \Maps(V,W)$ first as fixed points of a $G$-action, and then as invariant functions $\tilde f\in \Maps(V\times W^*, k)$. Thus, knowledge of invariant functions can be parlayed into knowledge of equivariant maps.

\section{Equivariance from invariants} \label{mechanics}

With the above imprecise philosophical discussion as a guide, Algorithm~\ref{alg:equivariant-from-invariant} shows how in practice to get from a description of invariant polynomials on $V\times W^*$, to equivariant polynomial (or smooth) maps $V\rightarrow W$. The technique given here is attributed to B. Malgrange; see \cite[Proposition~2.2]{ronga1975stabilite} where it is used to obtain the smooth equivariant maps, and \cite[Proposition~6.8]{schwarz1980lifting} where it is used to obtain holomorphic equivariant maps. Variants on this method are used to compute equivariant maps in \cite[Sections~2.1--2.2]{gatermann2007computer}, \cite[Section~3.12]{popov1994invariant},  \cite[Section~4.2.3]{derksen2015computational}, \cite[Section~4]{worfolk1994zeros}, and, in a machine learning context, in \cite[Section~2.2]{yarotsky2022universal}.

The goal of the algorithm is to provide a parametrization of equivariant maps. That said, the proof of correctness is constructive: as an ancillary benefit, it furnishes a method for taking an arbitrary equivariant map given by explicit polynomial expressions for the coordinates and expressing it in terms of this parametrization.

\begin{algorithm}[!hbt]
\caption{Malgrange's method for getting equivariant functions}\label{alg:equivariant-from-invariant}

\textbf{Input:} Bihomogeneous generators $f_1,\dots,f_m$ for $\RR[V\times W^*]^G$.

\begin{enumerate}
    \item Order the generators so that $f_1,\dots,f_r$ are of degree $0$ and $f_{r+1},\dots,f_s$ are of degree $1$ in $W^*$. Discard $f_{s+1},\dots,f_m$ (of higher degree in $W^*$).
    
    \item Choose a basis $e_1,\dots,e_d$ for $W$, and let $e_1^\top,\dots, e_d^\top$ be the dual basis, so an arbitrary element $\ell\in W^*$ can be written
    \[
    \ell = \sum_{i=1}^d \ell_i e_i^\top,
    \]
    and $\ell(e_i) = \ell_i$.
    
    \item \label{step:deriv} For $j=r+1,\dots, s$, and for $v\in V, \ell \in W^*$, let $F_j(v)$ be the total derivative of $f_j(v,\ell)$ with respect to $\ell\in W^*$, identified with an element of $W$ along the canonical isomorphism $W^{**}\cong W$; explicitly,
    \[
    F_j(v) := \sum_{i=1}^d \left( \frac{\partial}{\partial \ell_i} f_j(v,\ell)\right)e_i.
    \]
    Then each $F_j$ is a function $V\rightarrow W$.
\end{enumerate}

\textbf{Output:} Equivariant functions $F_{r+1},\dots,F_s$ from $V$ to $W$ such that any equivariant $f:V\rightarrow W$ can be written as
\[
f = \sum_{j=r+1}^s p_j(f_1,\dots,f_r)F_j.
\]

\textbf{Remark:} Because the $f_j(v,\ell)$'s are being differentiated with respect to variables in which they are linear, Step \ref{step:deriv} could alternatively have been stated: for each $j$, define $F_j$ as the vector in $W$ whose coefficients with respect to the $e_i$'s are  just the coefficients of the the $\ell_i$'s in $f_j(v,\ell)$. 
\end{algorithm}

We now exposit in detail Algorithm~\ref{alg:equivariant-from-invariant} and its proof of correctness, in the case where $f$ is a polynomial map; for simplicity we take $k=\RR$.  The argument is similar for smooth or holomorphic maps, except that one needs an additional theorem to arrive at the expression \eqref{eq:pullback} below. If $G$ is a compact Lie group, the needed theorem is proven in \cite{schwarz1975smooth} for smooth maps, and in  \cite{luna1976fonctions} for holomorphic maps over $\CC$.

We begin with linear representations $V$ and $W$ of a group $G$ over $\RR$. We take $W^*$ to be the contragredient representation to $W$, defined above. (If $G$ is compact, we can work in a coordinate system in which the action of $G$ on $W$ is orthogonal, and then we may ignore the distinction between $W$ and $W^*$, as discussed above in the case of $G=\mathrm{O}(d)$.) We suppose we have an explicit set $f_1,\dots,f_m$ of polynomials that generate the algebra of invariant polynomials on the vector space $V\times W^*$ (denoted as $\mathbb R[V\times W^*]^G$)---in other words, they have the property that {\em any} invariant polynomial can be written as a polynomial in these. We also assume they are {\em bihomogeneous}, i.e., independently homogeneous in $V$ and in $W^*$.  To reduce notational clutter we suppress the maps specifying the actions of $G$ on $V$ and $W$ (which were called $\varphi$ and $\psi$ in the previous section),  writing the image of $v\in V$ (respectively $w\in W$, $\ell \in W^*$) under the action of an element $g\in G$ as $gv$ (respectively $gw$, $g\ell$). We suppose $f_1,\dots,f_r$ are degree $0$ in $\ell$ (so they are functions of $v$ alone), $f_{r+1},\dots,f_s$ are degree $1$ in $\ell$, and $f_{s+1},\dots,f_m$ are degree $>1$ in $\ell$.

Now we consider an arbitrary $G$-equivariant polynomial function
\[
f : V \rightarrow W.
\]
We let $\ell \in W^*$ be arbitrary, and, as in the previous section, we construct the function
\begin{align*}
\tilde f: V \times W^* &\rightarrow \RR\\
(v,\ell)&\mapsto \ell(f(v)).
\end{align*}
Equivariance of $f:V\rightarrow W$ implies that $\tilde f$ is invariant:
\[
    \tilde f(gv, g\ell) = \ell\circ g^{-1}(f(gv))
    = \ell\circ g^{-1}(gf(v))
    = \ell(f(v)) = \tilde f(v,\ell).
\]
From the invariance of $\tilde f$, and the fact that $f_1,\dots,f_m$ generate the algebra of invariant polynomials on $V\times W^*$, we have an equality of the form
\begin{equation}\label{eq:pullback}
\tilde f(v,\ell) = P(f_1(v),\dots,f_m(v,\ell)),
\end{equation}
where $P\in \RR[X_1,\dots,X_m]$ is a polynomial. Note that $f_1,\dots,f_r$ do not depend on $\ell$, while $f_{r+1},\dots,f_m$ do.

We now fix $v\in V$ and take the total derivative $D_\ell$ of both sides of \eqref{eq:pullback} with respect to $\ell\in W^*$, viewed as an element of $W$.\footnote{In the background, we are using canonical isomorphisms to identify $W^*$ with all its tangent spaces, and $W^{**}$ with $W$.} Choosing dual bases $e_1,\dots,e_d$ for $W$ and $e_1^\top,\dots,e_d^\top$ for $W^*$, and writing $\ell = \sum \ell_ie_i^\top$, we can express the operator $D_\ell$ acting on a smooth function $F:W^*\rightarrow \RR$ explicitly by the formula
\[
D_\ell F = \sum_{i=1}^d \left(\frac{\partial}{\partial \ell_i}F\right) e_i.
\]
Applying $D_\ell$ to the left side of \eqref{eq:pullback}, we get
\begin{align*}
D_\ell \tilde f(v,\ell) &= \sum_{i=1}^d\left( \frac{\partial}{\partial \ell_i} \ell(f(v))\right)e_i \\
&= \sum_{i=1}^d \left(\frac{\partial}{\partial \ell_i} \left(\sum_{j=1}^n \ell_je_j^\top f(v)\right)\right) e_i\\
&= \sum_{i=1}^d \left(e_i^\top f(v)\right)e_i\\
&= f(v),
\end{align*}
so $D_\ell$ recovers $f$ from $\tilde f$. (Indeed, this was the point.) Meanwhile, applying $D_\ell$ to the right side of \eqref{eq:pullback}, writing $\partial_jP$ for the partial derivative of $P$ with respect to its $j$th argument, and using the chain rule, we get
\[
    D_\ell P(f_1,\dots,f_m) = \sum_{j=1}^m\partial_j P(f_1,\dots,f_m) D_\ell f_j.
\]
Combining these, we conclude
\begin{equation}\label{eq:express}
f(v) = \sum_{j=1}^m \partial_j P(f_1(v),\dots,f_m(v,\ell)) D_\ell f_j(v,\ell).
\end{equation}
Now we observe that $D_\ell f_j(v)=0$ if $j\leq r$, because in those cases $f_j(v)$ is constant with respect to $\ell$. But meanwhile, the left side of \eqref{eq:express} does not depend on $\ell$, and it follows the right side does not either; thus we can evaluate it at our favorite choice of $\ell$; we take $\ell = 0$. Upon doing this, $D_\ell f_j(v,\ell)|_{\ell=0}$ also becomes $0$ for $j>s$, because in these cases $f(v,\ell)$ is homogeneous of degree at least $2$ in $\ell$, so its partial derivatives with respect to the $\ell_i$ remain homogeneous degree at least $1$ in $\ell$, thus they vanish at $\ell = 0$. Meanwhile, $f_j(v,\ell)|_{\ell=0}$ itself vanishes for $j=r+1,\dots,m$, so that the $(r+1)$st to $m$th arguments of each $\partial_j P$ vanish. Abbreviating 
\[
\partial_j P(f_1(v),\dots,f_r(v),0,\dots,0)
\]
as $p_j(f_1(v),\dots,f_r(v))$, we may thus rewrite \eqref{eq:express} as
\[
f(v) = \sum_{j=r+1}^s p_j(f_1(v),\dots,f_r(v)) D_\ell f_j(v,\ell).
\]
Finally, we observe that, $f_j$ being linear homogeneous in $\ell$ for $j=r+1,\dots,s$, $D_\ell f_j(v,\ell)$ is degree $0$ in $\ell$, i.e., it does not depend on $\ell$. So we may call it $F_j(v)$ as in the algorithm, and we have finally expressed $f$ as the sum $\sum_{r+1}^s p_j(f_1,\dots,f_r)F_j$, as promised.

\section{Examples}\label{sec:examples}

In this section we apply Malgrange's method to parametrize equivariant functions in various examples. In all cases, we take a group $G$ of $d\times d$ matrices, equipped with its canonical action on $\RR^d$, and we are looking for equivariant maps from a tuple of vectors to a single vector.

\paragraph{The orthogonal group.}

For positive integers $d,n$, we parametrize maps $(\RR^d)^n \rightarrow\RR^d$ from $n$-tuples of $d$-vectors to a single $d$ vector, that are equivariant for the canonical action of $\mathrm{O}(d)$ on $\RR^d$. By the Riesz representation theorem, we can identify $\RR^d$ with  $(\RR^d)^*$ along the map $v \mapsto \langle v,\cdot\rangle$, where $\langle \cdot, \cdot \rangle$ is the standard dot product $\langle v,w\rangle = v^\top w$. Since $\mathrm{O}(d)$ preserves this product (by definition), this identification is equivariant with respect to the $\mathrm{O}(d)$-action, thus $(\RR^d)^*$ is isomorphic with $\RR^d$ as a representation of $\mathrm{O}(d)$. We may therefore ignore the difference between $W=\RR^d$ and $W^*$ in applying the algorithm.

Thus, we consider the ring of $\mathrm{O}(d)$-invariant polynomials on tuples 
\[
(v_1,\dots,v_n,\ell)\in (\RR^d)^n \times \RR^d=V\times W.
\]
We begin with bihomogeneous generators for this ring. By a classical theorem of Weyl \cite{weyl} known as the {\em First Fundamental Theorem for $\mathrm{O}(d)$}, they are the dot products $v_i^\top v_j$ for $1\leq i\leq j \leq n$; $v_i^\top\ell$ for $1\leq i\leq n$; and $\ell^\top\ell$. These are ordered by their degree in $W$ as in Step 1 of the algorithm; we discard $\ell^\top\ell$ as it is degree $>1$ in $W$.

We can work in the standard basis $e_1,\dots,e_d$ for $W = \RR^d$, and we have identified it with its dual, so Step 2 is done as well.

Applying Step 3, we take the generators of degree $1$ in $W$, which are
\[
f_1= v_1^\top \ell, \dots, f_n = v_n^\top \ell.
\]
Taking the total derivatives, we get
\begin{align*}
F_j(v_1,\dots,v_n) &= \sum_{i=1}^d \left(\frac{\partial}{\partial \ell_i} v_j^\top \ell\right) e_i\\
&= \sum_{i=1}^d (v_j)_ie_i\\
&= v_j,
\end{align*}
where $(v_j)_i$ denotes the $i$th coordinate of $v_j$. Thus the $F_j(v_1,\dots,v_n)$ yielded by the algorithm is nothing but projection to the $j$th input vector. Meanwhile, the $f_1(v),\dots,f_r(v)$ of the algorithm are the algebra generators $v_i^\top v_j$ of degree zero in $\ell$; thus the output of the algorithm is precisely the representation described in \eqref{eq:span-the-module} and the paragraph following.

\paragraph{The Lorentz and symplectic groups.}

If we replace $\mathrm{O}(d)$ with the Lorentz group $\mathrm{O}(1,d-1)$, or (in case $d$ is even) the symplectic group $\mathrm{Sp}(d)$, the entire discussion above can be copied verbatim, except with the standard dot product being replaced everywhere by the Minkowski product $v^\top \operatorname{diag}(-1,1,\dots,1) v$ in the former case, or the standard skew-symmetric bilinear form $v^\top J w$ (where $J$ is block diagonal with $2\times 2$ $\pi/2$-rotation matrices as blocks) in the latter. We also need to use these respective products in place of the standard dot product to identify $\RR^d$ equivariantly with its dual representation. The key point is that the invariant theory works the same way (\cite{weyl}; see \cite[Sec.~9.3]{popov1994invariant} for a concise modern treatment, noting that $O(d)$ and $O(1,d-1)$ have the same complexification).

\paragraph{The special orthogonal group.}

Now we consider $G=\mathrm{SO}(d)$. We can once again identify $\RR^d$ with its dual. However, this time, in Step 1, the list of bihomogeneous generators is longer \cite[Sec.~9.3]{popov1994invariant}: in addition to the dot products $v_i^\top v_j$ and $v_i^\top \ell$ (and $\ell^\top \ell$, which will be discarded), we have $d\times d$ determinants $\det(v_{i_1},\dots,v_{i_d})$ 
for $1\leq i_1\leq \dots \leq i_d\leq n$, and $
\det(v_{i_1},\dots,v_{i_{d-1}},\ell)$ 
for $1\leq i_1\leq \dots \leq i_{d-1}\leq n$. The former are of degree $0$ in $\ell$ while the latter are of degree $1$. Thus, the latter contribute to our list of $F_j$'s in Step 3, while the former figure in the arguments of the $p_j$'s. Carrying out Step 3 in this case, we find that 
\[
\sum_{i=1}^d \left(\frac{\partial}{\partial \ell_i}\det(v_{i_1},\dots,v_{i_{d-1}},\ell)\right) e_i
\]
is exactly the generalized cross product of the $d-1$ vectors $v_{i_1},\dots,v_{i_{d-1}}$. Thus we must add to the $F_j$'s a generalized cross-product for each $\binom{n}{d-1}$-subset of our input vectors; in the end the parametrization of equivariant maps looks like
\begin{align*}
f = &\sum_{i=1}^d p_i\left((v_i^\top v_j)_{\{i,j\}\in \binom{[n]}{2}}, \det(v|_S)_{S\in \binom{[n]}{d}}\right)v_i \\
&+ \sum_{S'\in \binom{[n]}{d-1}} p_{S'}\left((v_i^\top v_j)_{\{i,j\}\in \binom{[n]}{2}}, \det(v|_S)_{S\in \binom{[n]}{d}}\right)v_{S'},
\end{align*}
where $[n]:=\{1,\dots,n\}$, $\binom{[n]}{k}$ represents the set of $k$-subsets of $[n]$, $\det(v|_S)$ is shorthand for $\det(v_{i_1},\dots,v_{i_d})$ where $S=\{i_1,\dots,i_d\}$, and $v_{S'}$ is shorthand for the generalized cross product of the $d-1$ vectors $v_{i_1},\dots,v_{i_{d-1}}$ where $S'=\{i_1,\dots,i_{d-1}\}$.

\section{Discussion}

This article gives a gentle introduction to equivariant machine learning, and it explains how to parameterize equivariant maps $V\to W$ in two different ways. One requires knowledge of the irreducible representations and Clebsch-Gordan coefficients, and the other requires the knowledge of the generators of the space of invariant polynomials on $V\times W^*$. The main focus is on the latter, which is useful to design equivariant machine learning with respect to groups and actions where the invariants are known and are computationally tractable. This is not a panacea: sometimes the algebra of invariant polynomials is too large, or outright not known. Both these issues come up, for example, in the action of permutations on $n\times n$ symmetric matrices by conjugation, the relevant one for graph neural networks. The invariant ring has not been fully described as of this writing, except for small $n$, where the number of generators increases very rapidly with $n$; see \cite{thiery2000algebraic}.

From a practitioner's point of view, it is not yet clear which of these approaches will behave better in practice. We conjecture that Malgrange's construction applied to non-polynomial invariants such as \cite{olver2023invariants, dym2022low, cahill2022group}, as described at the end of Section~\ref{sec:inv-thy-for-ml}, is a promising direction for some applications, especially because some of them exhibit desirable stability properties.

\section{Acknowledgments} 
The authors thank Gerald Schwarz (Brandeis University) for introducing us to Malgrange's method. The authors also thank Jaume de Dios Pont (UCLA), David W. Hogg (NYU), Teresa Huang (JHU), and Peter Olver (UMN) for useful discussions. SV and BBS are partially supported by ONR N00014-22-1-2126. SV was also partially supported by the
NSF–Simons Research Collaboration on the Mathematical and Scientific Foundations of Deep
Learning (MoDL) (NSF DMS 2031985), the TRIPODS Institute for the Foundations of
Graph and Deep Learning at Johns Hopkins University, NSF CISE 2212457, and an AI2AI Amazon research award.

\bibliographystyle{abbrv}
\bibliography{ref} 
\end{document}